%% file: main.tex
\documentclass[sigconf,nonacm]{acmart}

\usepackage{booktabs} % For formal tables
\usepackage{listings}
\usepackage[utf8]{inputenc}
\usepackage{xcolor}
\usepackage{multirow}
\usepackage[binary-units=true]{siunitx}

\copyrightyear{2019}
\acmYear{2019} 
\setcopyright{acmlicensed}
\acmConference[IPSN '19]{The 18th International Conference on Information Processing in Sensor Networks (co-located with CPS-IoT Week 2019)}{April 16--18, 2019}{Montreal, QC, Canada}
\acmBooktitle{The 18th International Conference on Information Processing in Sensor Networks (co-located with CPS-IoT Week 2019) (IPSN '19), April 16--18, 2019, Montreal, QC, Canada}
\acmPrice{15.00}
\acmDOI{10.1145/3302506.3310390}
\acmISBN{978-1-4503-6284-9/19/04}

\begin{document}

%
% The "title" command has an optional parameter, allowing the author to define a "short title" to be used in page headers.
\title[Hazard Monitoring with Convolutional Neural Networks on the Edge]{Event-triggered Natural Hazard Monitoring with Convolutional Neural Networks on the Edge}

% \author{Matthias~Meyer, Timo~Farei-Campagna, Akos~Pasztor, Reto~Da~Forno, Tonio~Gsell, Samuel~Weber, Jan~Beutel, Lothar~Thiele}
% \affiliation{%
%   \institution{Computer Engineering and Networks Laboratory, ETH Zurich}
%   \city{Zurich}
%   \state{Switzerland}
% }
% \email{matthias.meyer@tik.ee.ethz.ch}

% \author{J\'erome~Faillettaz, Andreas~Vieli}
% \affiliation{
%     \institution{Department of Geography, University of Zurich}
%     \city{Zurich}
%     \state{Switzerland}
% }

\author{Matthias Meyer}
\orcid{1234-5678-9012}
\affiliation{%
  \institution{ETH Zurich}
}
\email{matthias.meyer@tik.ee.ethz.ch} 

\author{Timo Farei-Campagna}
\affiliation{%
  \institution{ETH Zurich}
}
\email{timofa@student.ethz.ch}

\author{Akos Pasztor}
\affiliation{%
  \institution{ETH Zurich}
}
\email{mail@akospasztor.com}

\author{Reto Da Forno}
\affiliation{%
  \institution{ETH Zurich}
}
\email{reto.daforno@tik.ee.ethz.ch}

\author{Tonio Gsell}
\affiliation{%
  \institution{ETH Zurich}
}
\email{tonio.gsell@tik.ee.ethz.ch}

\author{J\'erome Faillettaz}
\affiliation{
    \institution{University of Zurich}
}
\email{jerome.faillettaz@geo.uzh.ch}

\author{Andreas Vieli}
\affiliation{
    \institution{University of Zurich}
}
\email{andreas.vieli@geo.uzh.ch}

\author{Samuel Weber}
\affiliation{%
  \institution{ETH Zurich}
}
\email{samuel.weber@tik.ee.ethz.ch}

\author{Jan Beutel}
\orcid{0000-0003-0879-2455}
\affiliation{%
  \institution{ETH Zurich}
}
\email{janbeutel@ethz.ch}

\author{Lothar Thiele}
\affiliation{%
  \institution{ETH Zurich}
}
\email{lothar.thiele@tik.ee.ethz.ch}

\renewcommand{\shortauthors}{M. Meyer et al.}

\begin{abstract}
In natural hazard warning systems fast decision making is vital to avoid catastrophes. Decision making at the edge of a wireless sensor network promises fast response times but is limited by the availability of energy, data transfer speed, processing and memory constraints. In this work we present a realization of a wireless sensor network for hazard monitoring based on an array of event-triggered single-channel micro-seismic sensors with advanced signal processing and characterization capabilities based on a novel co-detection technique. On the one hand we leverage an ultra-low power, threshold-triggering circuit paired with on-demand digital signal acquisition capable of extracting relevant information exactly and efficiently at times when it matters most and consequentially not wasting precious resources when nothing can be observed. On the other hand we utilize machine-learning-based classification implemented on low-power, off-the-shelf microcontrollers to avoid false positive warnings and to actively identify humans in hazard zones. The sensors' response time and memory requirement is substantially improved by quantizing and pipelining the inference of a convolutional neural network. In this way, convolutional neural networks that would not run unmodified on a memory constrained device can be executed in real-time and at scale on low-power embedded devices. A field study with our system is running on the rockfall scarp of the Matterhorn H\"ornligrat at \SI{3500}{\meter}~a.s.l. since 08/2018.
\end{abstract}

 \begin{CCSXML}
 <ccs2012>
 <concept>
<concept_id>10002951.10003317.10003347.10003356</concept_id>
<concept_desc>Information systems~Clustering and classification</concept_desc>
<concept_significance>300</concept_significance>
</concept>
<concept>
<concept_id>10003033.10003083.10003090.10011643.10011650.10011663</concept_id>
<concept_desc>Networks~Wireless mesh networks</concept_desc>
<concept_significance>300</concept_significance>
</concept>
<concept>
<concept_id>10003033.10003106.10003112.10003238</concept_id>
<concept_desc>Networks~Sensor networks</concept_desc>
<concept_significance>300</concept_significance>
</concept>
<concept>
<concept_id>10010147.10010257.10010293.10010294</concept_id>
<concept_desc>Computing methodologies~Neural networks</concept_desc>
<concept_significance>500</concept_significance>
</concept>
<concept>
<concept_id>10010583.10010588.10010595</concept_id>
<concept_desc>Hardware~Sensor applications and deployments</concept_desc>
<concept_significance>100</concept_significance>
</concept>
<concept>
<concept_id>10010583.10010588.10010596</concept_id>
<concept_desc>Hardware~Sensor devices and platforms</concept_desc>
<concept_significance>100</concept_significance>
</concept>
</ccs2012>
\end{CCSXML}

\ccsdesc[300]{Information systems~Clustering and classification}
\ccsdesc[500]{Computing methodologies~Neural networks}
\ccsdesc[300]{Hardware~Sensor applications and deployments}
\ccsdesc[300]{Hardware~Sensor devices and platforms}
\ccsdesc[100]{Networks~Wireless mesh networks}
\ccsdesc[300]{Networks~Sensor networks}

\keywords{Natural Hazard Warning System, Wireless Sensing Platform, Embedded Convolutional Neural Network, Machine Learning, On-device Classification }

%
% This command processes the author and affiliation and title information and builds
% the first part of the formatted document.
\maketitle

\input{content}

\bibliographystyle{ACM-Reference-Format}
\bibliography{IPSN2019.bib}

\end{document}

%% file: content.tex
% Definition of results
%% Training results
\newcommand{\noninqfscore}{0.9693}
\newcommand{\inqfscore}{0.9779}
\newcommand{\noninqerrorrate}{0.0329}
\newcommand{\inqerrorrate}{0.0240}

%% Latency results
\newcommand{\inftimeshort}{1.08~s}
\newcommand{\inftimelong}{9.55~s}
\newcommand{\inftimetdp}{0.44~s}

%% Memory results
\newcommand{\memoryshort}{245.76~kB}
\newcommand{\memorylong}{2375.68~kB}
\newcommand{\memorytdp}{85.6~kB }
\newcommand{\memoryimprovement}{2.87}

\section{Introduction}
In the following we present a scenario where it is mandatory to perform complex decision making on the edge of a distributed information processing system and show based on a case-study how the approach can be embedded into a wireless sensor network architecture and what performance can be expected.

Natural disasters happen infrequently and for mitigation efforts fast reaction times relative to these rarely occurring events are important, especially where critical infrastructure or even human casualties are at stake \cite{gladeLandslideHazardRisk2005}. In alpine regions where human habitats including settlements and infrastructure are threatened by rockfalls and other gravity-driven slope failures, wireless sensor networks can act as natural warning systems~\cite{intrieriDesignImplementationLandslide2012}. They have the flexibility to be deployed in locations that are logistically difficult or dangerous to access, for example an active rockfall scarp. Therefore it is important that these systems run autonomously for long periods of time~\cite{girardCustomAcousticEmission2012, Beutel_IEEE2009}. Unfortunately, in many cases the close proximity of warning systems to the human habitat has negative implications as noise originating from infrastructure or anthropogenic activities may impact the capabilities and accuracy of a warning system and therefore must be accounted for. In this paper we demonstrate how human noise can be classified, quantified and removed from microseismic signals using an implementation of a convolutional neural network optimized for embedded devices.

Traditionally, continuous, high-resolution data acquisition is used to monitor microseismic signals emanating from structural fatigue \cite{amitranoSeismicPrecursoryPatterns2005,occhienaAnalysisMicroseismicSignals2012}. These methods are powerful in capturing natural hazard with respect to process understanding as well as hazard warning. However, they suffer severely that in periods of no or little activity continuous high-rate signal amplification and sampling does not provide an information gain while still consuming energy. In addition, these methods scale unfavorably due to the large amount of data produced \cite{werner-allenFidelityYieldVolcano2006}. Overall increasing the number of sensor nodes leads to improved detection capability but is clearly limited by the available transmission bandwidth if data is to be processed centrally. One way to reduce the network utilization is to make the sensors themselves more intelligent by shifting the knowledge generation process from the centralized backend closer to the signal sources, e.g.~\cite{girardCustomAcousticEmission2012}. A novel approach based on a coupled fibre-bundle model exists~\cite{faillettazCodetectionAcousticEmissions2016a}. It registers precursory patterns of catastrophic events with the help of many threshold triggered sensors and a reduced set of explanatory variables and has recently been tested in a pre-study \cite{Faillettaz_nhess2018}. Thus, a system that is optimized to calculate explanatory variables directly on the sensor reduces the logging and transmission cost and thus allows to react with high reactivity on the detection of catastrophic events~\cite{Beutel_Date2011}. 

But such an extreme reduction in information content comes at a price: false positives and the inability to characterize events further. While the first has an impact on correct analysis and network performance metrics, the latter is of importance to react adequately on the detection of a disaster. For example if humans are present in a hazard zone they should be warned and a search and rescue mission should be dispatched immediately. Correct and timely information here is of utmost importance to maximize success and avoid expensive interventions on false alarms. Under the constraints given, on-device classification provides a mean to identify humans on-location without the requirement to transmit all sensory data through the network.

We present two aspects: (i) a system architecture for natural hazard monitoring using seismic sensors (geophones) that is designed for detection of precursory rockfall patterns based on the theory of co-detection and (ii) a concept for accurate on-sensor classification of event-based seismic signals to reduce false positives and enhance information by identifying humans in the signal using machine learning techniques. 

We evaluate our network and system architecture in two scenarios. In the first scenario we present an outdoor, wide-area sensing system presently deployed in a high alpine natural hazard environment at the Matterhorn Hörnligrat field site at \SI{3500}{\meter}~a.s.l., Zermatt, Switzerland. We demonstrate the functionality of our system architecture and evaluate its longevity when using event-based data acquisition. The second scenario is a laboratory experiment using an openly available microseismic dataset~\cite{meyerMicroSeismicImageDataset2018} to demonstrate the feasibility of on-device classification of mountaineers using convolutional neural networks within the wireless sensor architecture. We focus on advanced methods to reduce the memory requirements and latency of an embedded convolutional neural network classifier.

In this context the paper contains the following contributions
\begin{itemize}
    \item A realization of a wireless, event-triggered single-channel microseismic sensor system featuring low-power consumption, fast wake-up time and on-device signal processing and characterization capabilities. The system is realized using the Dual-Processor Platform (DPP) hardware design template~\cite{suttonBoltStatefulProcessor2015} and a slightly adapted version of the open-source protocol implementation of the event-based Low-Power Wireless Bus (eLWB)~\cite{suttonDesignResponsiveEnergyefficient2017a}.
    \item An implementation of a convolutional neural network for seismic event classification on low-power embedded devices using network quantization.
    \item A sophisticated buffering concept for pipelined inference of a convolutional neural network to relax memory requirements and to decrease latency.
\end{itemize} 
 
Complementary material related to this paper, such as code, is provided online \cite{meyerComplementaryMaterials2019}. 

\section{Related Work}

\textbf{Rockfall Detection using Seismic Sensors: }
Seismic precursory patterns before rockfalls have been investigated for several field sites \cite{senfauteMicroseismicPrecursoryCracks2009,spillmannMicroseismicInvestigationUnstable2007, amitranoSeismicPrecursoryPatterns2005}. These studies are based on microseismic measurements with a portable data logger. Wireless sensor networks have been introduced to cover a larger area while removing the requirement of data retrieval \cite{werner-allenMonitoringVolcanicEruptions2005,colomberoMicroseismicityUnstableRock2018,occhienaAnalysisMicroseismicSignals2012}. They either provide the option for remote data download or transmit short, event-triggered segments. Unlike in our study, event triggering is done in the digital domain which means that the acquisition system is constantly on. 

\textbf{Acoustic Event Detection: }
Artificial neural networks have been applied to acoustic event classification \cite{takahashiDeepConvolutionalNeural2016,hersheyCNNArchitecturesLargeScale2016,cakirPolyphonicSoundEvent2015a,cakirConvolutionalRecurrentNeural2017} which includes among others footstep detection. Also footstep detection and person identification using geophones has been studied before \cite{anchalUREDTUnsupervisedLearning2018, panIndoorPersonIdentification2015,lamRobustOccupantDetection2016}, however only in experiments in a controlled environment, not on embedded  devices or using additional structural information.
Artificial neural networks have been recently applied to seismic event detection \cite{paitzNeuralNetworkNoise2018}. Especially convolutional neural networks have achieved good accuracies \cite{perolConvolutionalNeuralNetwork2018,meyerSystematicIdentificationExternal2019}. 

\textbf{Artificial Neural Networks on Embedded Devices: }
Many studies focus on additional accelerators \cite{andriYodaNNArchitectureUltralow2018,hegdeUCNNExploitingComputational2018,chenEyerissEnergyEfficientReconfigurable2017} for convolutional neural networks. This approach requires dedicated hardware. Studies on mobile platforms \cite{wuQuantizedConvolutionalNeural2016} and wearables \cite{mathurDeepEyeResourceEfficient2017} exist but require a more powerful hardware architecture. A prominent work for low-power embedded devices focuses on keyword spotting \cite{zhangHelloEdgeKeyword2017a} on a slightly more powerful Cortex-M7 than used in our study. A theoretic strategy for low-memory convolutional neural networks as been proposed in \cite{binasLowmemoryConvolutionalNeural2018} which focuses on an incremental depth-first processing idea that resembles our approach. However, they neither focus on sequential data nor on the implementation with a specific buffering system.

\section{System Design}
\begin{figure}
    \centering
    \includegraphics[width=\columnwidth]{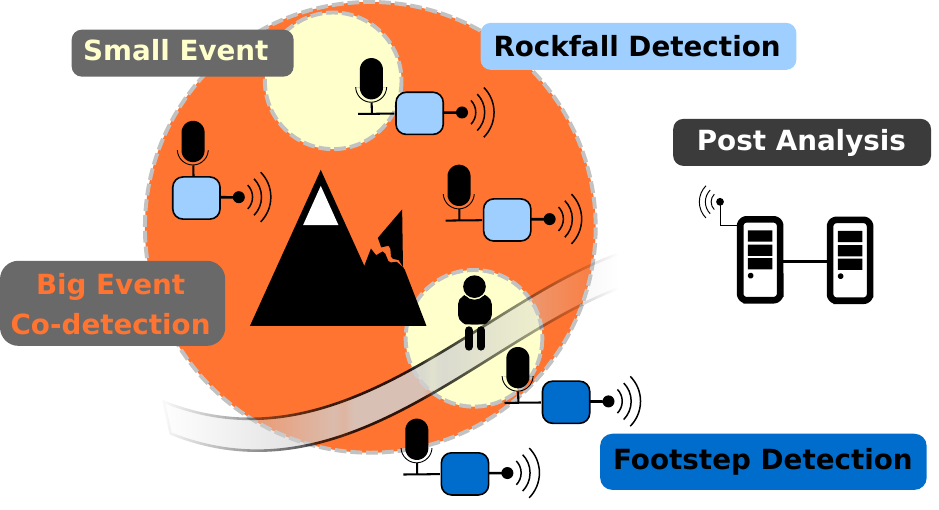}
    \caption{Conceptual illustration of a wireless sensor network for natural hazard monitoring based on the principle of co-detection~\cite{faillettazCodetectionAcousticEmissions2016a}. Multiple seismic sensors are deployed in a hazardous area. The sensor nodes feature the same hardware, a microseismic sensor (geophone), processing, storage and wireless communication subsystems and are able to detect and classify events based on threshold-triggered input signal. If a sensor detects an event it can determine if it originates from a human in the hazard zone or not and possibly trigger an alert. If not it only sends the event information trough a wireless low-power network. A basestation collects the information from all sensors for centralized data gathering and further analysis using post-processing methods. Temporal correlation of detected events, e.g. when  multiple sensors register an event within the same time window, make it possible to identify the precursors of large mass failures based on the theory of co-detection~\cite{faillettazCodetectionAcousticEmissions2016a}.} 
    \label{fig:codetection_system}
\end{figure}

Figure~\ref{fig:codetection_system} illustrates the overall concept. A wireless sensor network consisting of multiple microseismic sensor nodes is deployed in an area where rockfall occurs. The system can be partitioned into sensor nodes that are only used as rockfall detectors (light blue) and sensor nodes that additionally can classify footsteps (dark blue). The two node types have different requirements which will be briefly outlined in the following.

\subsection{Rockfall Detection by Co-detection of Seismic Events}\label{sec:codetection}
The following describes the principle of detecting precursors of rockfall patterns~\cite{faillettazCodetectionAcousticEmissions2016a} with threshold-triggered geophone sensors. Multiple geophones are deployed on the rock surface as illustrated in Figure~\ref{fig:codetection_system}. If rockfall stimulates a seismic event either due to fracturing/detaching or due to impact, different sensors may register the emerging signal depending on their location relative to the event source~\cite{Weber_JGR2018}. A high amplitude input signal registered at a single sensor can have two causes: Either a large event occurred at distance or a small event occurred in close proximity to the sensor. A co-detection exists if multiple sensors register an event quasi-simultaneously, which allows to distinguish between the two aforementioned possibilities. Furthermore, as lab experiments have shown~\cite{faillettazCodetectionAcousticEmissions2016a}, consecutive co-detections of events can be used to identify rockfall precursors and thus facilitates natural hazard early warning capabilities. Fundamental for this principle is the requirement of many sensors to perform co-detection as well as to cover a large enough area with a sensor cluster~\cite{Faillettaz_nhess2018}. The data acquisition can be reduced to recording the exact timestamp when the signal exceeds a certain threshold, i.e. capturing events only. While this detection can be implemented very efficiently in hardware using an analog comparator circuit further analysis using signal processing techniques require a digitizer and processing unit that is typically put to sleep when not in use. A predictable system behaviour and a precise time synchronization between all system components and all nodes is important be able to put co-detected events into context and quantify the underlying processes. A similar system based on much higher frequency acoustic emission signals has been implemented successfully  using the Dual Processor Platform (DPP) architecture \cite{suttonBoltStatefulProcessor2015} and the event-based Low-Power Wireless Bus (eLWB)~\cite{suttonDesignResponsiveEnergyefficient2017a}. In this work, we adopt these openly available system components to realize an outdoor, wide-area sensing system~\cite{Pasztor_MSc2018}, evaluate and demonstrate its applicability to perform co-detection of precursory rockfall patterns based on low-frequency microseismic signals.

\begin{figure}
    \centering
    \includegraphics[width=\columnwidth]{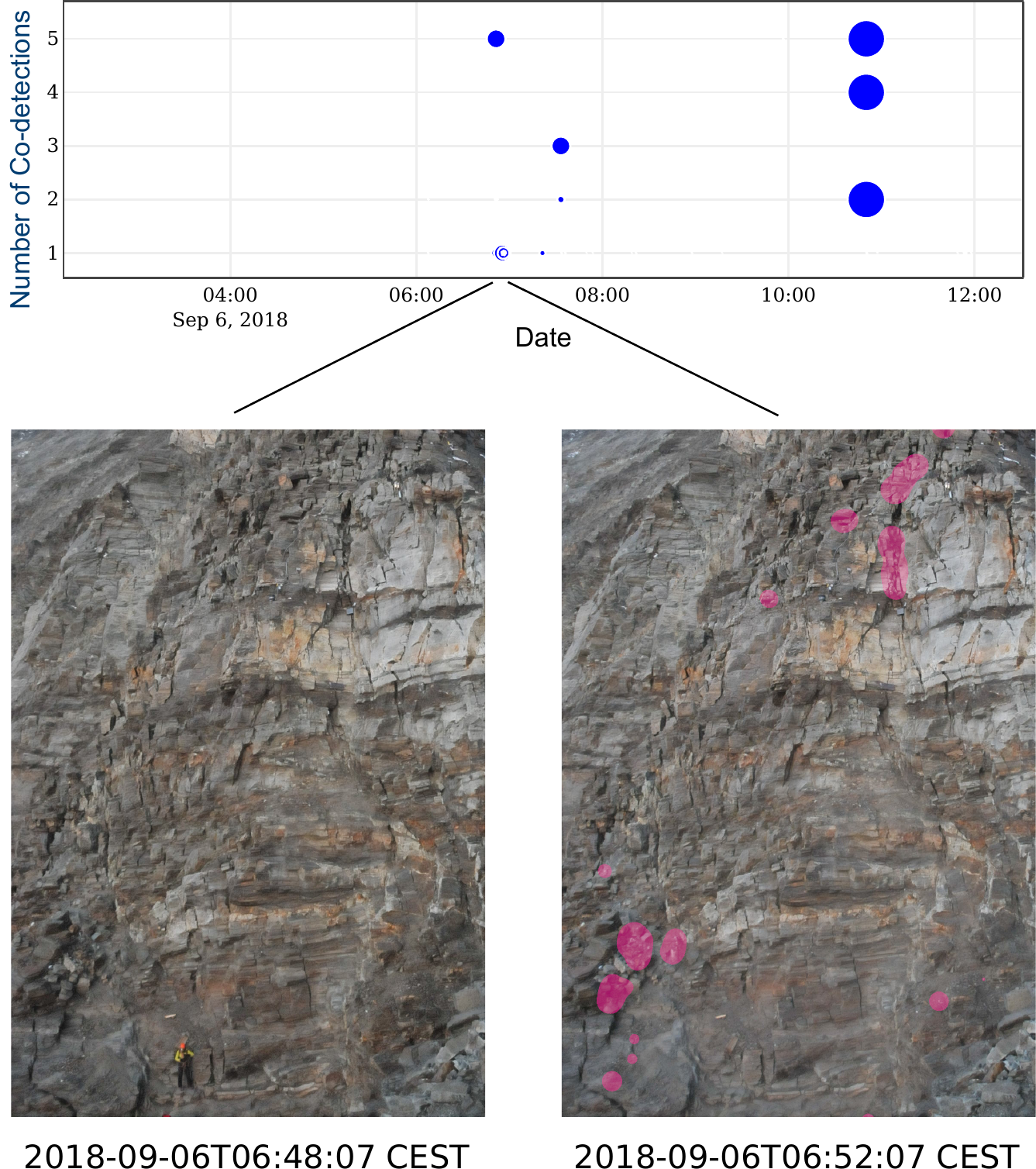} 
    \caption{An example of a rockfall event during the testing phase of the wireless sensor network at the Matterhorn Hörnligrat field-site~\cite{Weber_JGR2018}. Illustrated is data from the monitoring system and two images (before and after significant rockfall) obtained from a remotely controlled high-resolution camera. The top plot shows how a cluster of sensors (see Figure \ref{fig:instrumentation}) co-detected an event over time. The point size indicates the maximum peak amplitude detected in each co-detection within a 0.5 second time window whereas the vertical axis denotes how many sensors triggered within this window. Marked in pink on the right image are areas impacted by rockfall identified by comparing the two images. The mountaineer visible in the lower left corner of the left image is in the danger zone with a number of significant impacts visible in immediate vicinity (pink). Local reports confirmed that no one was harmed in this specific incident.} 
    \label{fig:rockfall}
\end{figure} 

\subsection{Classification with Time Distributed Processing}
The co-detection concept allows to reduce false positives, e.g. caused by anthropogenic activity like humans walking by. However, to identify whether a human is present on-site is impossible by just using the reduced set of information transmitted by the event-triggered sensors, i.e. the timestamps of detected events. Since transmitting the raw sensor data of each detected event in real-time for many sensor is infeasible due to bandwidth and energy limitations an approach using on-device classification is advocated. Here, several challenges need to be addressed. Multiple footstep detectors using geophones have been proposed \cite{anchalUREDTUnsupervisedLearning2018,panIndoorPersonIdentification2015} but have not been shown to distinguish well between footsteps and seismic events \cite{meyerSystematicIdentificationExternal2019} or require further structural information \cite{lamRobustOccupantDetection2016}. Convolutional neural networks have shown to be good signal processing tools for classification of acoustic \cite{hersheyCNNArchitecturesLargeScale2016} as well as seismic sources \cite{perolConvolutionalNeuralNetwork2018}. In contrast to other neural network types, such as MLP or LSTM, several convolutional neural network architectures for the special case of seismic event detection have been explored \cite{perolConvolutionalNeuralNetwork2018,meyerSystematicIdentificationExternal2019}. Thus, this work focuses on optimizing and implementing an existing CNN-based classifier with known good performance \cite{meyerSystematicIdentificationExternal2019} to perform well on embedded devices. 
On the downside, convolutional neural networks have a high memory demand, high memory access rates and a high processing demand. Typical commercially available low-power embedded devices are equipped with two types of memories, static random-access memory (SRAM) and flash memory. On low-power devices the impact of memory usage on the energy efficiency is significant and space in energy-efficient memory structures (SRAM) is limited. However, the inference of a convolutional neural network requires a significant amount of memory to perform the computations, specifically for storing intermediate results and the network parameters. Non-volatile memory, such as flash memory, is typically used to store the parameters of the neural network but the number of read accesses to this type of memory should be minimized since the energy consumption is typically about 6x as high as reading from SRAM \cite{verykiosSelectivePoliciesEfficient2018}. As a consequence the amount of memory accesses required for loading parameters should be reduced, for example by binarization of the network \cite{mcdanelEmbeddedBinarizedNeural2017}. However this approach comes with a drop in accuracy of about 10\%. In our work we apply incremental network quantization \cite{zhouIncrementalNetworkQuantization2017}, which does not suffer from a reduced accuracy while reducing the network parameter's memory requirement.  

For storing intermediate results SRAM is the most energy-efficient memory. However, the intermediate results of state-of-the-art convolutional neural networks do not fit into SRAM. Additionally, convolutional neural networks suffer from a high latency because of the high number of operations required to perform a classification. 

In this work we present a novel method to pipeline the computations which relaxes the memory requirements significantly and allows to compute a convolutional neural network in SRAM only while providing a low latency. In the following we call this concept time distributed processing.
 
\section{Wireless Sensing Platform} \label{sec:sensing_platform}
\begin{figure*}
    \centering
    \includegraphics[width=16cm]{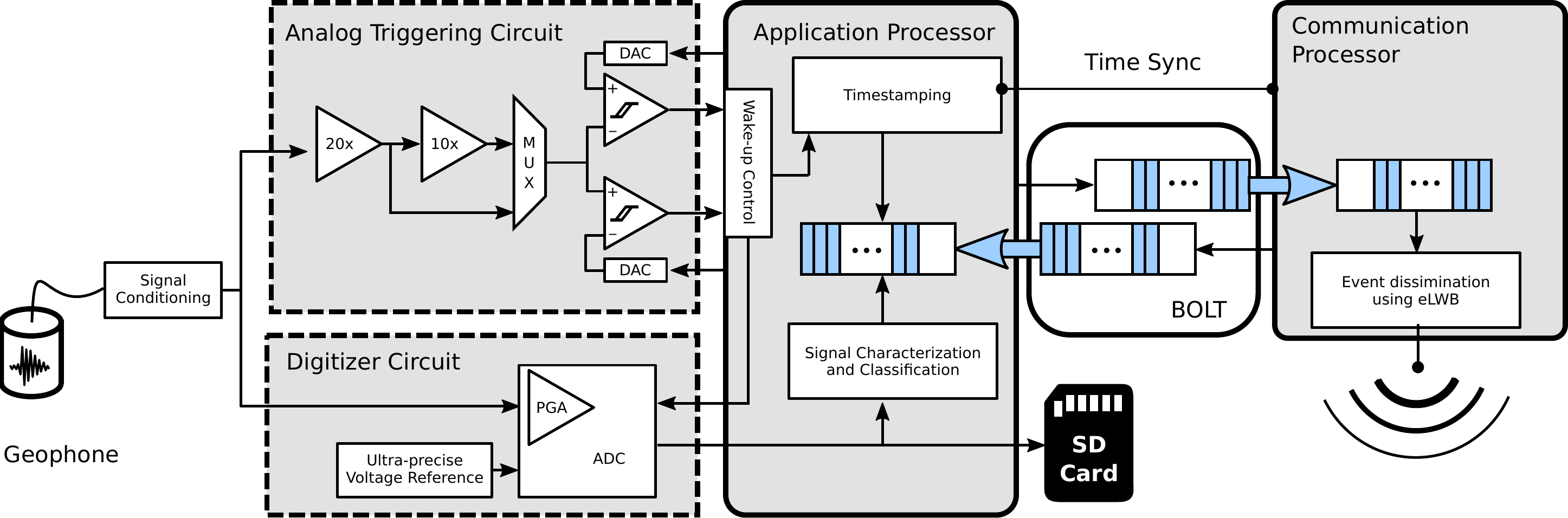}
    \caption{System diagram of the event-triggered microseismic sensing platform based on the Dual Processor Platform (DPP) architecture template~\cite{suttonBoltStatefulProcessor2015, suttonDesignResponsiveEnergyefficient2017a}. The analog triggering circuit as well as the BOLT interface is always powered whereas all other components can make use of low-power operating modes independently.} 
    \label{fig:dpp}
\end{figure*}

The wireless event-triggered microseismic sensing platform presented in this paper is depicted in Figure~\ref{fig:dpp} and consists of one single-axis geophone sensor, an analog triggering circuit, a digitizer circuit, an application processor, a communication processors integrated using the BOLT state-full processor interconnect~\cite{suttonBoltStatefulProcessor2015}. We are using a single-axis, omni-tilt geophone sensor since it can be used over a large range of the inclination angles, a characteristic usually not available on multi-axial geophone sensors that require an accurate and level placement over the whole measurement duration. The geophone signal is conditioned and fed to the analog triggering circuit and digitizer circuit. The analog triggering circuit provides the application processor with an interrupt signal if the geophone signal is higher or lower than a given threshold. The application processor will timestamp the detected event and then enable the digitizer system to sample the geophone signal for a pre-defined duration. The processing system is based on the Dual Processor Platform (DPP) partitioning and decoupling the sensing application and the communication onto dedicated processing resources. The interconnect on DPP is realized using BOLT~\cite{suttonBoltStatefulProcessor2015}, an ultra-low power processor state-full interconnect which features bi-directional, asynchronous message transfer and predictable run-time behaviour. The communication subsystem is based on an IEEE 802.15.4-compatible transceiver (MSP CC430) running eLWB~\cite{suttonDesignResponsiveEnergyefficient2017a}. Further details on the design of the system architecture can be found in~\cite{Pasztor_MSc2018} as well as a pre-study using a wired setup in \cite{Faillettaz_nhess2018}.

\subsection{Analog Triggering Subsystem}\label{sec:analog_triggering}
The circuit must be capable of amplifying the geophone sensor (SM-6 14Hz Omni-tilt Geophone, ION Geophysical Corporation) signal and comparing it to a predefined threshold trigger. This part of the system is always active, therefore it is most sensitive with regards to power consumption. All other system components can be duty-cycled but the trigger must remain powered at all times. An evaluation of different implementation variants showed the superiority of a fully discrete external solution (140.5, 115.2, 22.9 uA respectively) among variants using OPAMP, DAC and comparator circuits internal to the application processor (32-bit ARM-Cortex-M4, STM32L496VG, 80MHz core clock, 3 uA current drain in STOP 2 mode with full RAM retention with RTC on; 1MB flash, 320~kB continuous SRAM), a mix with external and internal components or all-external components respectively. The final design incorporates a dual-sided trigger with individual threshold set-points and a variable amplification (20x and 200x) using MAX5532 12-bit DACs and MAX9019 comparators. The input signal is biased to half the rail voltage and the upper and lower thresholds can be selected between $0 - V_{sys}/2$ and $V_{sys}/2 - V_{sys}$ respectively. Although (in theory) one single-sided trigger should be sufficient, we deliberately chose to implement a dual-trigger system (triggering both on a rising and falling first edge of the seismic signal) to be able to have more degrees of freedom and stronger control over the trigger settings chosen. The overhead for the bipolar trigger system relative to the whole systems power figures in it's different operating modes is negligible (see Section \ref{sec:evaluation}). 

\subsection{Digitizer Subsystem}\label{sec:dpp_dsp}
Upon detection of a threshold crossing of the incoming sensor signal the application processor is woken up from an external interrupt and a timestamp of this event is stored. Subsequently the data acquisition system, a 24-bit delta-sigma ADC with high SNR and built-in Programmable Gain Amplifier and low noise, high-precision voltage reference (MAX11214 ADC) is powered on and initialized. It samples the geophone signal at 1~ksps and stores data in SD card storage  until the signal remains below the trigger threshold values for a preconfigured duration (post-trigger interval). For this purpose all successive threshold crossings of the sensor signal (the interrupts) are monitored. After ADC sampling has completed the ADC is switched off and all data describing the detected event (event timestamp, positive/negative threshold trigger counts, event duration, peak amplitude, position of peak amplitude) are assembled into a data packet that is queued for transmission over the wireless network along with further health and debug data packets. Using this data, rockfall detection by means of co-detection as described earlier in section \ref{sec:codetection} can commence using only very lightweight data traffic while the full waveform data is available for further processing and event classification as presented later in section \ref{sec:event_classification}.

\subsection{Wireless Communication System}
The communication system is based on the TI CC430 system-on-chip running an adapted version of the event-based Low-Power Wireless Bus (eLWB) \cite{suttonDesignResponsiveEnergyefficient2017a} based on Glossy. This protocol provides low-latency and energy-efficiency for event-triggered data dissemination using interference-based flooding. Since the protocol was specifically designed to be triggered by ultra-low power wake-up circuits it is optimally suited for our application. We use the openly available code\footnote{https://github.com/ETHZ-TEC/LWB} with adaptations specific to our platform and the data to be transferred.

\subsection{Application Integration}
The Dual Processor Platform (DPP) philosophy using the BOLT state-full processor interconnect~\cite{suttonBoltStatefulProcessor2015} builds on the paradigm of separation of concerns, shielding different system components and run-time functionality from each other for as much as possible. As a side effect this partitioning allows for easy integration and adaption to new applications and/or specifications by allowing to work on communication and application separately. Also, by using well-defined and strongly de-coupled interfaces application re-use is facilitated. In BOLT two queues implemented on non-volatile memory form a strictly asynchronous interface between two processors with guaranteed maximum access times. The obvious drawback of this strict de-coupling however is, that all interaction between the two processors is message based and incurs different end-to-end delays depending on queue fill and access patterns. Therefore tight time synchronization is not readily available. For this purpose a dedicated sync signal is routed between interrupt capable IOs of the two processors. In this way both the decoupling of the two application contexts for sensing and communication as well as tight time sync for accurate timestamping of the detected events based on the network-wide high-precision time sync of eLWB can be achieved~\cite{Pasztor_MSc2018}. 

Apart from the event-triggered geophone sensing platform design and its system integration the main contribution of this work is to demonstrate and evaluate a blueprint method for on-board characterization and classification of detected events using neural networks and machine learning techniques. The key techniques and challenges encountered are discussed in the following two sections.

\section{Event Classification}\label{sec:event_classification}
\begin{figure}
    \centering
    \includegraphics[width=\columnwidth]{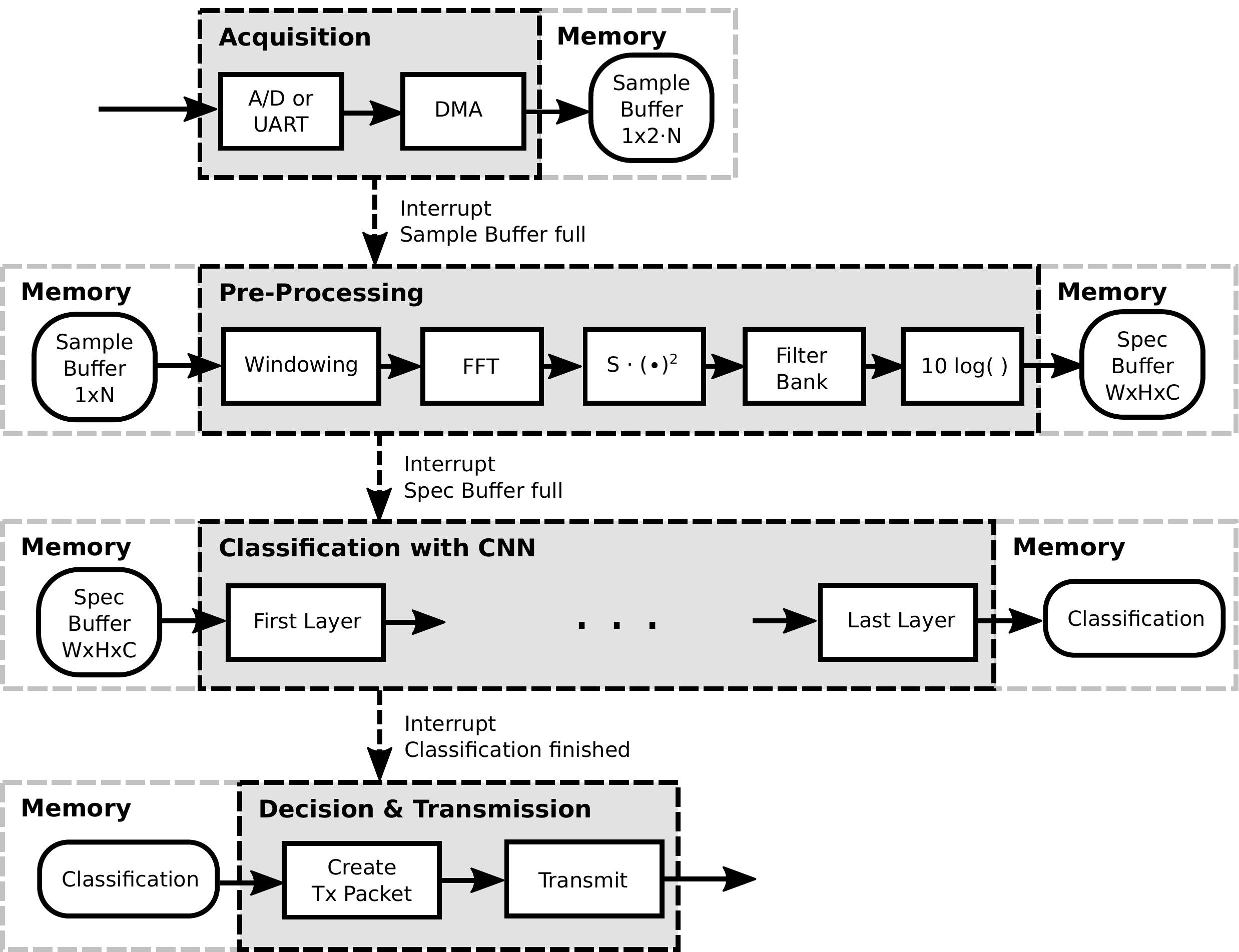}
    \caption{Signal processing pipeline for event classification on the experimental platform. The processing pipeline is subdivided into data acquisition, pre-processing, classification with a convolutional neural network (CNN) and decision an transmission. Data sharing between tasks is realized via buffers. Whenever the sample buffer is filled the pre-processing executes and fills one column of the spec buffer. When $W$ columns are filled the classification task is executed. The result of the classification can be transmitted.} 
    \label{fig:signal_processing_classification}
\end{figure}

The training of an event classifier for a new field site is always affected by the cold-start problem: Little knowledge about the data is available at the time of initial deployment but this knowledge is required to train a classifier. Moreover, the size and diversity of the dataset is critical for training a good classifier requiring a long time period of samples, for example sensor signals observed over multiple seasons or in different weather conditions.
To mitigate this issue we perform a preliminary feasibility study on a dataset with similar characteristics to our application scenario. In this way we can assess the energy and storage requirements which are necessary to deploy nodes at a later stage using on-device classification. The neural network can later be retrained easily when an extensive dataset is acquired with the integrated system.
The processing pipeline for event classification is illustrated in Figure~\ref{fig:signal_processing_classification} and consists of data acquisition, pre-processing, classification with a convolutional neural network and result transmission. For evaluation purposes we use a development board with the same micro-controller as used for the application processor on the sensor nodes presented earlier in Section~\ref{sec:sensing_platform}. The micro-controller's UART module allows to input data and output results with the flexibility to feed different experimental data for development, debug and performance assessments. In the final design the digitizer frontend with the analog-digital-converter and the network data packet generation feeding result data over BOLT to the communication subsystem will replace these UART modules.

\subsection{Training Dataset}
The dataset used is an openly accessible microseismic dataset captured at Matterhorn Hörnligrat \cite{meyerMicroSeismicImageDataset2018}. One sample of the dataset consists of a two-minute microseismic recording and a camera image. Both data types coincide in time. The sample's label indicates whether mountaineers are present on the image or not. The dataset also contains additional data structures, such as a list of event timestamps. To be able to use our system model with the dataset two changes have to be applied to the dataset. The seismic sensor used in the reference dataset is a three-axial seismometer (Lennartz LE-3Dlite MkIII). Since our sensor nodes are equipped with a single-axis geophone sensor only a single channel is available as classifier input. Thus, we only use the vertical component of the three-axis dataset for training and testing. The characteristics of the geophone and the seismometer~\cite{meyerMicroSeismicImageDataset2018} are comparable for the signals of concern in this application. We apply an amplitude triggering algorithm to the two-minute signals and retrieve 12.8 second long event segments to which we assign the same labels as the respective two-minute segments. We set the threshold such that the number of events per two-minute segment is similar in quantity to the event timestamps provided by the dataset. These event segments are then used for training and evaluation. We use the same split for training and test set as defined in the dataset.

The processing pipeline illustrated in Figure~\ref{fig:signal_processing_classification} transforms the digitized geophone signals into a time-frequency representation which the convolutional neural network uses for classification. The dataset samples are acquired over the UART interface to ensure a repeatable experimental setup and subsequently stored in memory using efficient Direct Memory Access (DMA). The samples are transferred via UART using a sampling frequency of 1000 samples per second which is a comparable rate as in the sensing platform presented in Section~\ref{sec:sensing_platform}. When the sample buffer is filled an interrupt triggers the processing task. We perform strided segmentation and segment the signal with a segment size of N=1024 and a stride of 512 using a double buffer of size $2N$.

\subsection{Pre-Processing}\label{sec:preprocessing}
The pre-processing on the embedded systems is equal to the processing used to train the neural network. It is designed to be efficiently implemented using a Fast Fourier Transform (FFT). Other techniques for audio or seismic classification work directly on the time-domain signal \cite{paitzNeuralNetworkNoise2018}, however in that case the convolutional neural network tends to learn a time-frequency representation \cite{sainathLearningSpeechFrontend2015}. By using a FFT the efficiency of its implementation can be exploited in contrast to implementing a filter bank with convolutional filters. The pre-processing task takes the sample buffer as input, multiplies it with a Tukey window ($\alpha = 0.25$) and performs the FFT. The magnitude of the FFT is squared, scaled and transformed using a filterbank. The filterbank maps the FFT bins to 64 bins and thus reduces the data to be processed and stored in a later stage of the signal processing pipeline. Consecutive log compression creates a distribution of values which is more suitable for the convolutional neural network \cite{choiComparisonAudioSignal2017}. With an input segment size of 12.8 seconds the size of the time-frequency representation is Time x Frequency x Channels (T x F x C) = 24x64x1.

\subsection{Convolutional Neural Network}\label{sec:neural_net}
We use a neural network for classification of mountaineers that is openly accessible \cite{meyerCodeClassifierTraining2018} and which has already been structurally optimized for a reduced parameter set and few computations. It consists of multiple convolutional layers with rectified linear unit (ReLU) activation and zero padding to match input and output size. Moreover, dropout is used to reduce overfitting.  In contrast to \cite{meyerCodeClassifierTraining2018} we do not use Batch Normalization layers because we found it to have negligible impact on the test accuracy in our experiment.  Our implementation is illustrated in Table~\ref{tab:network}.

For evaluation of the neural network we will use error rate and the F1 score which is defined as

\begin{equation}
\text{F1\ score} = \frac{2 \cdot \text{true\ positive}}{2 \cdot \text{true\ positive} +  \text{false\ negative} +  \text{false\ positive}}
\end{equation}

To prevent overfitting the neural network is all-convolutional  \citep{springenbergStrivingSimplicityAll2014} and dropout \citep{srivastavaDropoutSimpleWay2014} is used. Training is performed using Tensorflow \cite{tensorflow2015-whitepaper} and Keras \cite{cholletKeras2015}. It is accomplished by using 90\% of the training set to train while a random 10\% of the training set is used for validation and never used during training. The number of epochs is set to 100. For each epoch the F1 Score is calculated on the validation set and the epoch with the best F1 score is selected. The test accuracy is determined independently on the test set.

\subsection{Implementation Challenges on Embedded Devices}
To implemented the neural network on an embedded device further optimizations are required. The first problem is the storage required for the parameters of the network. The number of parameters is 38,403 which requires 153.6~kB of flash memory using 32-bit values. It is possible to store this amount in flash memory but read accesses to flash should be minimized due to the higher power consumption in comparison to reading from SRAM \cite{verykiosSelectivePoliciesEfficient2018}. We therefore apply Incremental Network Quantization \cite{zhouIncrementalNetworkQuantization2017} which quantizes the parameters to power-of-two values in an iterative weight partition and quantization process. Due to quantization the parameters can be stored as 8 bit integer values and the storage for the parameters of the convolutional neural network is reduced by a factor of 4 without loss in classification accuracy. 

The second problem is the size of the intermediate results. The largest intermediate result of the convolutional neural network is calculated in layer $C_0$. To calculate layer $C_1$ the output from layer $C_0$ and additional space to store the output of $C_1$ is required. With 32-bit values the memory requirement in our case is \memoryshort~kB which is too large to fit into the SRAM of most micro-controller units. Of course, provisioning this amount of memory would be possible, e.g. using external memory but due to the increase in silicon, access times and energy footprint alternative methods need to be sought for. Since external DRAM is not a suitable solution either we present a method which allows to execute the convolutional neural network using only SRAM and a reduced memory footprint in the following section.

\begin{table}
    \centering
    \begin{tabular}{ccccc}
        Name & Type & Kernel & Stride & Input size \\
        \hline
        $C_0$ & Conv2D + ReLU & 3x3 & 1 & 24 x 64 x 1\\
        $C_1$ & Conv2D + ReLU & 3x3 & 2 & 24 x 64 x 32\\
        $D_0$ & Dropout & - & - & 12 x 32 x 32 \\
        $C_2$ & Conv2D + ReLU & 3x3 & 1 & 12 x 32 x 32\\
        $C_3$ & Conv2D + ReLU & 3x3 & 2 & 12 x 32 x 32\\
        $D_1$ & Dropout & - & - & 6 x 16 x 32 \\
        $C_4$ & Conv2D + ReLU & 3x3 & 1 & 6 x 16 x 32\\
        $C_5$ & Conv2D + ReLU & 1x1 & 1 & 6 x 16 x 32\\
        $D_2$ & Dropout & - & - &  6 x 16 x 32\\
        $C_6$ & Conv2D + ReLU & 1x1 & 1 & 6 x 16 x 32\\
        $A_f$ & Average (Frequency) & 1x16 & 1 & 6 x 16 x 1 \\
        $A_t$ & Average (Time) &  6x1 & 1 & 6 x 1 x 1 \\
        $C_7$ & Conv2D + Sigmoid & 1x1 & 1 & 1 x 1 x 1 \\
        $O$ & Output & - & - & 1 \\
    \end{tabular}
    \caption{The structure of the convolutional neural network using 2D convolutional layers (Conv2D) with Rectified Linear Units (ReLU) and dropout layers to reduce overfitting. Number of parameters 38,403. }
    \label{tab:network}
\end{table}

\section{Memory Footprint Reduction Approach: Time Distributed Processing} 
\begin{figure}
    \centering
    \includegraphics[width=\columnwidth]{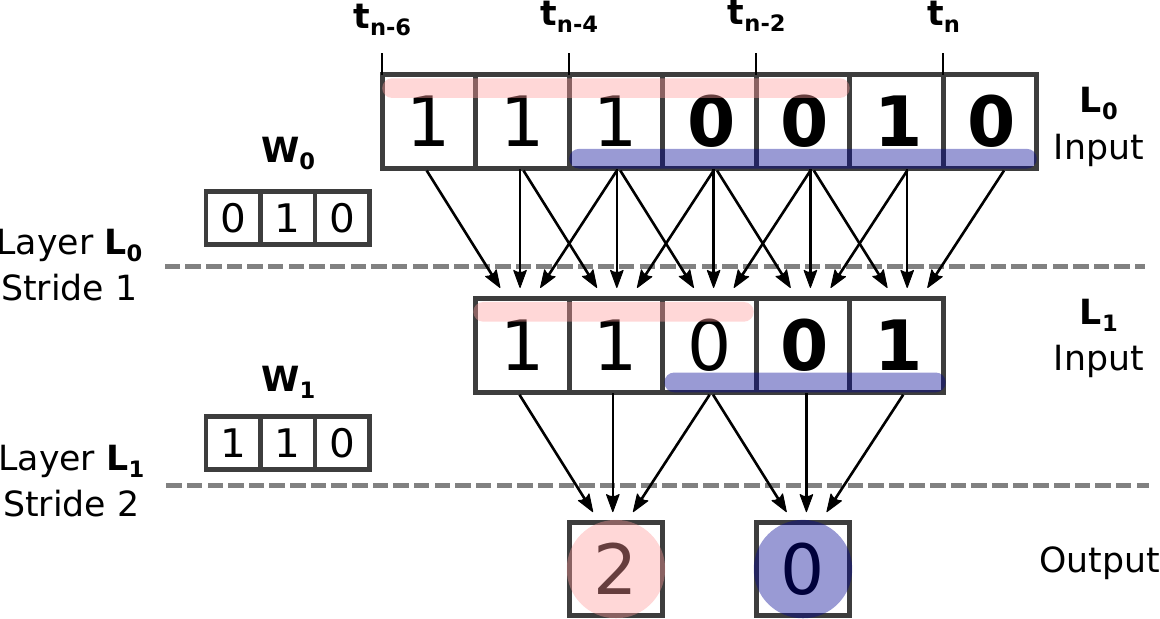}
    \caption{The receptive field of the output values is illustrated in red and in blue and grows with the number of layers. The values are calculated layer by layer by convolving the layer input with the respective kernel $w$.} 
    \label{fig:normal_inference}
\end{figure}

In this section we present a method to reduce the memory footprint requirement of the convolutional neural network. We will explain this concept with a simple example of a 1D convolutional neural network as illustrated in Figure~\ref{fig:normal_inference}. The network consists of two convolutional layers with a 3~x~1 weight kernel each and strides of 1 and 2, respectively. For illustration purpose we ignore the non-linearity and the bias which are usually part of a convolutional layer. Typically, the network is calculated layer by layer. The input is convolved with the first layer's parameters and the first layer's output is convolved with the seconds layer's parameters, which requires the intermediate outputs to be simultaneously in memory for the time of execution. 

In contrast to this approach we will focus on calculating the output values step by step. Illustrated in red and blue are the respective receptive fields of the second layers' outputs, meaning all values of the input buffer and first layer's output that affect the final output. We first calculate the red output, then we calculate the blue output. This idea is similar to the depth-first approach described in \cite{binasLowmemoryConvolutionalNeural2018}. However, additionally we will optimize for the temporal characteristics of our input data with a sophisticated buffering system. 

When calculating the red output we already calculated one intermediate result required for the blue output (the point where the local receptive fields in the layer 1 output overlap). If we want to calculate the blue output we see that we only need two values from the first layer's output which have never been calculated before (highlighted in bold). By following the receptive field of these two values we find they depend on four values from the input buffer, among which are two values which have not been used before and two which have been used for calculating the red output. Effectively, this means that we can calculate the output of the network by a combination of new input values, intermediate results and buffered values.

\begin{figure}
    \centering
    \includegraphics[width=8cm]{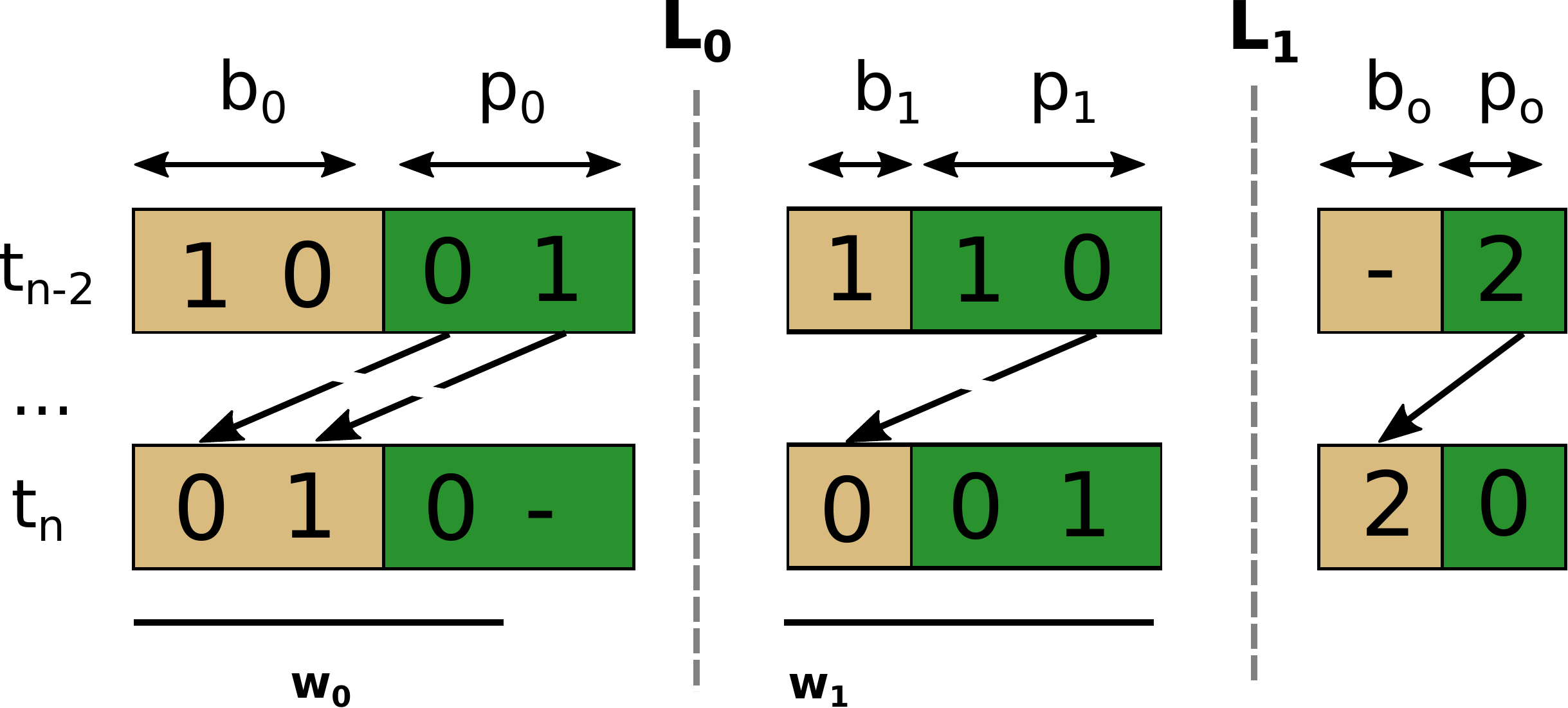}
    \caption{Illustration of the buffering concept: Each input of a layer consists of one buffer (yellow) and one processing window (green). Shown are two timesteps $t_{n-2}$ and $t_n$. Each timestep a new input is placed in the processing buffer of layer $L_0$. Subsequently, the layers perform convolutions followed by a left-shift of the processing window by two values.} 
    \label{fig:buffering_inference}
\end{figure}

Figure~\ref{fig:buffering_inference} illustrates this buffering concept for the same example. We use the following nomenclature: $p_i$ is the number of new input values for layer $L_i$; $b_i$ is the number of buffered input values for layer $L_i$; $p_o$ and $b_o$ are the respective output values. We call the array of size $p_i$ the processing window of $L_i$ and the array of size $b_i$ the buffer of layer $L_i$. The value $s_i$ is the stride for layer $L_i$. The figure presents the last two time steps $t_{n-2}$ and $t_n$, which results in calculating the blue value in Figure~\ref{fig:normal_inference}. Based on Figure~\ref{fig:normal_inference} we set $p_0=p_1=2$, $p_o=1$ and $b_0=2$,$b_1=b_o=1$. We now apply the convolutional layer as before but on the reduced input window. Obviously, we can now shift the input step by step into the buffering system and calculate the output of the convolutional neural network. For each step we acquire two new input samples and discard two older input samples. This makes the buffering system ideal for time series data as is the case in our application scenario. The memory requirement in Figure~\ref{fig:buffering_inference} compared to the case in Figure~\ref{fig:normal_inference} is reduced by 35\% and the memory requirement for each step is constant. The following section addresses the questions how this general concept is transferred to our application specific convolutional neural network introduced in  Section~\ref{sec:neural_net} and how the correct size of each layers' buffers is determined.

\subsection{Buffer System Design}
Closely examining the inputs of our convolutional neural network in Table~\ref{tab:network} we see that it consists of 24 timesteps of a 64 value vector. The buffering concept can be expanded to this case if we assume $b_0$x64x1 and $p_0$x64x1 input buffers. Similarly it can be expanded to multiple feature maps (in our case 32) for the intermediate buffers. Before determining the actual buffer sizes we need to determine up to which layer the buffering concept is applicable. The last three layers of the network are average pooling operations and scaling. The output of the time average pooling layer $A_t$ is influenced by every value of the 24x64x1 network input, which means that after this layer the buffering concept cannot be applied anymore. We can however calculate for each step one of the 6 input values to $A_t$ independently and place them in a buffer. Therefore we can consider all layers up to $A_t$ for the buffer system.

The buffer system is defined by the processing window size $p_i$ and buffer size $b_i$. The $p_i$ values can be calculated for each layer by considering the convolutional layers up to $A_t$.

$$ p_i = \prod_{l=i}^{L-1} s_l $$

where $L$ is the number of convolutional layers up to $A_t$ and $s_l$ is the stride of layer $C_l$. Note that we ignore the dropout layers for this calculation since they have no influence on inference.

Instead of deriving $b_i$ analytically we followed a systematic approach for each layer. Similar to the approach in the above example we choose the correct $b_i$ based on the layers $p_i$, $p_{i+1}$, kernel size and stride.

\begin{figure*}
    \centering
    \includegraphics[width=\textwidth]{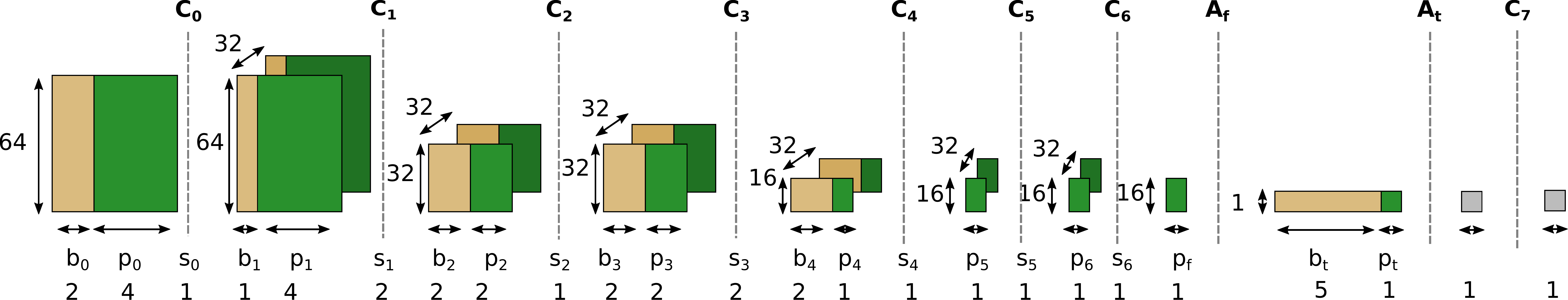}
    \caption{Illustrates the buffering architecture for the network from Table~\ref{tab:network} with input size of 4x64x1. The input and output of the network are depicted as well as the intermediate results of each layer. $C_0$ to $C_7$ are the convolutional layers. $A_f$ and $A_t$ are the average pooling layers for frequency and time, respectively.} 
    \label{fig:buffering}
\end{figure*}

Finally we can construct our buffering system for the network presented in \ref{tab:network}. The result is illustrated in \ref{fig:buffering}. As discussed before, the number of samples on the frequency axis remains unchanged as well as the number of feature maps.

Using this technique we are able to reduce the memory requirements from \memoryshort \ to \memorytdp \ for our example application which constitutes a reduction by a factor of $\sim$\memoryimprovement \ and as a result allows an efficient implementation on a resource-limited and sufficiently low-power embedded processing device.

\section{Result and Evaluation}\label{sec:evaluation}
In the following we will present our findings, evaluate our system design and demonstrate the advantages of time distributed processing.  The dataset \cite{meyerMicroSeismicImageDataset2018} is used to assess the performance of the mountaineer classifier. These results are then used in combination with performance data from our field deployment to estimate the lifetime of a sensor node equipped with the event classifier integrated onto the platform's embedded application processor.
We do not provide a qualitative evaluation of the rockfall detection system since a labelled dataset including every rockfall during a substantial monitoring period would be required but currently something like this does not exist (worldwide).

\subsection{Field Site Experiments}
\begin{figure}
    \centering
    \includegraphics[width=\columnwidth]{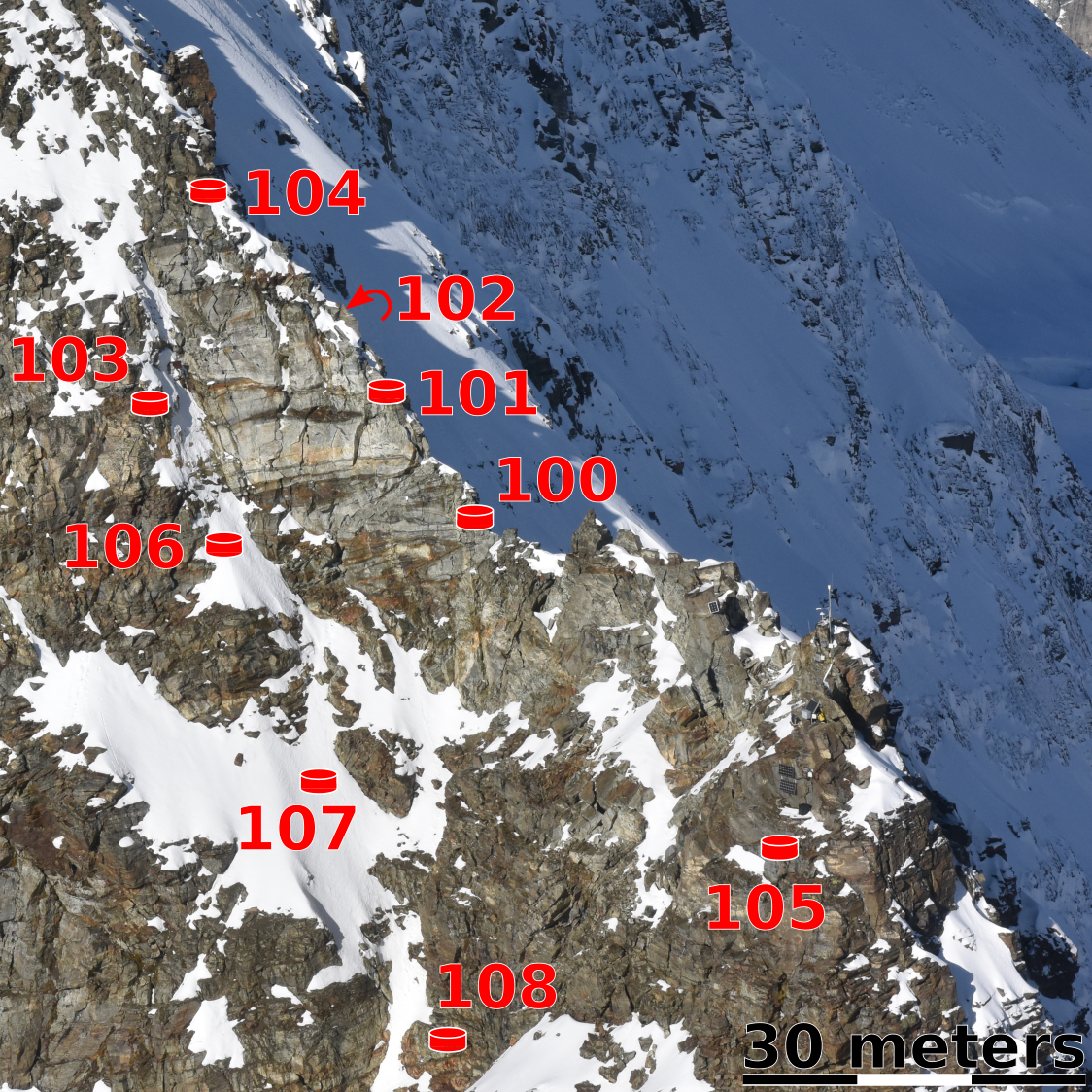}
    \caption{Matterhorn Hörnligrat Field site overview: Shown are the locations of the nine wireless geophone sensor nodes. The base station to collect data and transmit it to a data backend is located behind the little tower visible at the upper lefthand corner of the image. The detachment scarp visible is located above a frequently used climbing route.} 
    \label{fig:instrumentation}
\end{figure}

Nine geophone nodes were deployed in steep, fractured bedrock permafrost on the Matterhorn Hörnligrat field site~\cite{Weber_JGR2018}, a site prone to frequent rockfall hazards to evaluate the system characteristics and the suitability for co-detection of rockfall events. The locations of the event-triggered sensor nodes are depicted in Figure~\ref{fig:instrumentation}. The system is located right around a frequently used climbing route and continuously operating since mid-August 2018. The data from the site is fed into a web-based data portal and openly available\footnote{http://data.permasense.ch/}.

\subsection{Wireless Sensing Platform Evaluation}
The system characteristics of the wireless sensing platform presented in Section~\ref{sec:sensing_platform} is evaluated in terms of responsiveness and energy efficiency. 
Lab measurements (see \ref{tab:lab_measurments}) have shown that the 20x gain single amplification stage requires only \SI{29}{\micro\ampere}@\SI{3.0}{\volt} with the 200x gain dual amplification stage consuming \SI{49.9}{\micro\ampere}@\SI{3.0}{\volt}. In combination with the other components of the trigger frontend that are continuously running the system requires a current of \SI{35}{\micro\ampere}@\SI{3.0}{\volt} in sleep mode when using the single stage amplification. The active current with ADC operating and application processor running was measured to be \SI{35}{\milli\ampere}.

\begin{table}
    \centering
    \begin{tabular}{p{5cm}c}
        Measurements                                                            & Value [mA]\\
        \hline
        Active current CC430/eLWB                                               & 28        \\
        Sleep current CC430/eLWB                                                & 0.005     \\
        \hline
        Active current for geophone sensor, frontend and application processor  & 35        \\
        Sleep current geophone                                                  & 0.035     \\
        \hline
    \end{tabular}
    \caption{Lab measurements of the power performance of the event-triggered microseismic sensor platform at different characteristic operating modes.}
    \label{tab:lab_measurments}
\end{table}

The wake-up time based on an event trigger is important for the acquisition of event waveform data. Since the data acquisition on the ADC is not running continuously in order to save power, no pre-trigger samples are available.Moreover, the delay between threshold-based triggering and the ADC acquiring a first sample will result in data loss with respect to the event signal acquired. We measured a wake-up time from sleep mode of the processor to the acquisition of the first sample on the ADC of only 2.62~ms, which means that on average we loose approximately 3 ADC samples when using a sampling rate of 1~ksps. For most seismic data acquisition systems 1~ksps can be assumed as being a very high sampling rate with a typical value being only 250~sps. Therefore we conclude that using our system architecture this delay is not of significance for the given application.

Time synchronization is a crucial design criteria for our triggered sensing application and especially for using co-detection. Implementing this application using the eLWB protocol based on a synchronous network operating paradigm and Glossy flooding on the lower layer, we are able to achieve time synchronization in realistic operating conditions across a network of tens of nodes within \SI{200}{\micro\second}~\cite{suttonBLITZNetworkArchitecture2017}. 

\begin{table}
    \centering
    \begin{tabular}{p{5cm}c}
         Statistic &  Value\\
        \hline
        Number of Sensors & 9 \\
        Days in Field & 43 \\
        Total Sensor On-Time (h) & 28.227 \\
        Total Number of Events & 29040 \\
        Mean Number of Events per Hour & 28.14 \\
        Mean Event Length (s) & 3.5 \\
        \hline
        Mean Daily Acquisition Per Sensor (kB) & 788 \\
        Mean On-Time Per Hour Per Sensor (s) & 10.941 \\
        Mean Events per Hour Per Sensor & 3.127 \\
        Sensor duty cycle (\%) & 0.304 \\
        \hline
        Average current CC430 (mA) & 0.845 \\
        Average current geophone (mA) & 0.141 \\
        Average current total (mA) & 0.986 \\
        \hline
        Energy per day (mAh) & 23.667 \\
        Battery capacity (Ah) & 13 \\
        Estimated lifetime (days) & 549 \\
    \end{tabular}
    \caption{The first 43 days of the test phase at our field site have been used to collect statistics about the system behaviour. These statistics in combination with lab measurements have been used to estimate the average current of a sensor node and its expected lifetime: $\sim$1.5~years using a standard D-size lithium battery (SAFT LSH-20).}
    \label{tab:statistics} 
\end{table}

\begin{figure}
    \centering
    \includegraphics[width=\columnwidth]{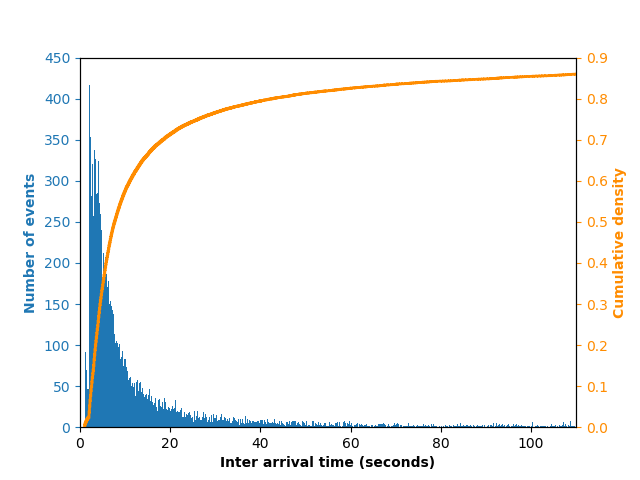}
    \caption{The histogram (blue) shows the absolute frequency of the inter-arrival time for $0.1\,\textrm{s}$ bins indicating that a large fraction of events occur in bursts with small inter-arrival times. The cumulative density is visualized in orange.} 
    % 14\% cut off
    \label{fig:interarrival}
\end{figure}

For further evaluation of the wireless sensor platform we use data from the first 43 days of the testing phase of our deployment. As can be seen from the statistics presented in Table~\ref{tab:statistics}, the mean number of events per hour is approximately 28. This value takes into account all events from all sensors and shows that the activity in the network is rather low and low-power performance in sleep mode is most important for this specific application. By looking at the histogram of inter-arrival times of all events in Figure~\ref{fig:interarrival} we can see that most events are occurring in bursts, having small inter-arrival times below 20 seconds. However, the cumulative density indicates that approximately 15\% of inter-arrival times are larger than 100 seconds and thus that there are as well long silent periods, which is also indicated by an inter-arrival time mean of $\sim$1044~seconds. This finding supports our choice of providing the system with an event-triggered sensing system since we can save energy during these silent periods. 

The activity statistics can further be used to estimate the energy consumption of our system. With an average event length of 3.5~seconds and the event count per hour per sensor we can estimate the duty cycle of the sensor to be 0.304\%. Additionally, using the measurements from the lab for sleep current and active current we can calculate the average current of one sensor node to be 0.986~mA. Using a battery of 13~Ah we can estimate the lifetime of one sensor node to be 549~days.

Using the event triggered sensor around 788~kB are recorded on average every day. Continuous sampling with one of our sensors would produce approx. 259 MB of data per day and sensor, which is an increase by a factor of $\sim$328. Similarly, assuming the geophone is continuously on and the communication processor sends packets as before, the estimated lifetime would be reduced by $\sim$95\% to 15~days excluding effects of higher bandwidth requirements and network congestion that would inevitably occur when building on the same wireless subsystem.

\subsection{Time Distributed Processing Evaluation}
The test results for the convolutional neural network from Section~\ref{sec:event_classification} are depicted in Table~\ref{tab:classifier_results}.  The error rate on the test set for the non-quantized network is \noninqerrorrate \ and the F1 Score \noninqfscore. These results are slightly worse than the test error rate after quantization, which is \inqerrorrate \ and the F1 score is \inqfscore. The effect that quantization improves the test error rate has been also been observed by the authors of the algorithm.

To underline the benefit of time distributed processing we compare the memory requirement and the latency of CNN inference for two scenarios: (i) inference of a 12.8 seconds window (which is the length of the window used for network training) and (ii) inference of an approximately 2 minute long window (the maximum window length we can train with the given dataset).

\begin{table}
    \centering
    \begin{tabular}{ccc}
         & Non-INQ & INQ \\
        \hline
        Top F1 Score  & \noninqfscore & \inqfscore\\
        Top Error rate & \noninqerrorrate & \inqerrorrate\\
    \end{tabular}
    \caption{Results for classifiers evaluated on the test set. The classifier trained with incremental network quantization (INQ) performs better than the classifier trained regularly. }
    \label{tab:classifier_results}
\end{table}

\subsubsection{Latency}
The convolutional neural network must process a time-frequency representation of a specific size, in our case 24x64x1 values, to perform one classification. The acquisition time is the time it takes to sample and pre-process the data to generate this time-frequency representation. The inference time is the time it takes to perform the calculation of the convolutional neural network. The latency is the sum of the acquisition time and the inference time. Table~\ref{tab:latency} shows the inference time for the two different input lengths. As can be seen, the larger the input the larger the gain of using time distributed processing since the inference time is constant for time distributed processing.
Through pipelining the calculation of the convolutional neural network we are more responsive since given the pre-processing details from Section~\ref{sec:preprocessing} and a $p_0$ of 4 the acquisition time is 2.56 seconds which is longer than our inference time of \inftimetdp.

\subsubsection{Memory requirement}
The memory requirement for the regular layer by layer inference is defined by the biggest intermediate result, since we can reuse the buffers used for calculation. For time distributed processing the memory requirement is defined by the size of the buffers, which are not reused in our implementation. 
When we compare the two input lengths from before we can see in Table~\ref{tab:latency} that on the one hand we are able to reduce the memory requirement for inference to only \memorytdp and on the other hand we see that it is independent of the input length. This independence on the input length is another key benefit of time distributed processing.

We used incremental network quantization to reduce our parameters by a factor of 4. The parameters consume 153.612~kB without network quantization and 38.403~kB with network quantization.
The sum of weight size and buffer size for intermediate results consumes together around 120~kB which fits in the 320~kB SRAM. Consequently, the parameters can be loaded once into SRAM and the processing can benefit from faster memory access and less energy consumption for reading the parameters. 

\begin{table}
    \centering
    \begin{tabular}{c|cc} 
        & w/o TDP & TDP  \\
        \hline
        Inference time (12.8~s window) & \inftimeshort & \inftimetdp \\
        Inference time (119.3~s window) & \inftimelong & \inftimetdp \\
        \hline
        Memory (12.8~s window) & \memoryshort & \memorytdp \\
        Memory (119.3~s window) & \memorylong & \memorytdp \\
    \end{tabular}
    \caption{Memory requirement and latency for inference of a convolutional neural network with and without our approach of time distributed processing (TDP). Shown are the values for two different input lengths.}
    \label{tab:latency} 
\end{table}

\subsubsection{Energy analysis}
Measurements on our evaluation platform show that the CPU duty cycle during inference is 10\%. The measured active current is 15~mA. However, we do not expect an increase in energy consumption when running the neural network on the wireless geophone node since the geophone node does not use any sleep mode during sampling. The application CPU and other components are always-on during sampling. Therefore the major impact is the longer sampling time of 12.8~s plus inference time of 0.44~s. Our estimation for the lifetime of the event-triggered, mountaineer-detection node is 422 days.

\section{Conclusion}
In this work we have presented a wireless sensor network architecture for the detection of rockfall events using event-triggered microseismic sensors and a method to perform machine-learning-based classification of events on low-power, memory-constraint devices. The system architecture has been designed and optimized for an application in natural hazard warning system providing additional information about human presence in a hazard zone. Our study shows that the lifetime of the system can be significantly extended through optimization for energy-efficiency by using analog triggering and on-device signal characterization. The resulting lifetime is $\sim$37x longer than when using continuous sampling while providing the relevant information for rockfall detection by co-detection of seismic events. In this way we demonstrate based on a real system implementation that information about imminent rockfall and potential hazard to human life including real-time warnings can be acquired in an efficient way, with latencies of only few seconds and in scenarios of realistic scale. Furthermore we demonstrate the performance of this system in a long-term field experiment in a realistic setting on the Matterhorn Hörnligrat, Zermatt, Switzerland.

Moreover, we demonstrate the feasibility of implementing a convolutional neural network for characterization of seismic signals using the example of footstep detection on a low-power microprocessor with a limited SRAM of only 320~kB. By using a network quantization we are able to reduce the parameter's memory requirement by a factor of 4. Additionally, we present a strategy to pipeline a convolutional neural network for temporal data such that we can significantly reduce the inference-time and the inference-memory requirement by a factor of $\sim$\memoryimprovement \ and keep them constant independent of the temporal size of the convolutional neural network input.

\section{Acknowledgement}
The work presented in this paper is part of the X-Sense 2 project financed by nano-tera.ch (ref. no. 530659). We would like to thank the the PermaSense and TEC team for integration support. We thank Lukas Cavigelli, Francesco Conti, Mahdi Hajibabaei for insightful discussions and the anonymous reviewers for valuable feedback.